\documentclass[runningheads]{llncs}
\usepackage{graphicx}
\usepackage{url}

\usepackage{xcolor}
\usepackage{bmpsize}
\usepackage{moreverb,url}
\usepackage{pgfplots}
\usepackage{pgfplotstable}
\usepackage{sfmath}
\usepackage{textcomp}
\usepackage{pdflscape}

\usepackage[textsize=scriptsize]{todonotes}

\usepackage[pdftex,%
pdfauthor={Simon M. Haller-Seeber (Simon.Haller-Seeber@uibk.ac.at), Thomas Gatterer, Patrick Hofmann,
Christopher Kelter, Thomas Auer, Michael Felderer},%
pdftitle={Software Testing, AI and Robotics (STAIR) Learning Lab},%
pdfkeywords={digital twin, educational robotics, physical computing, artificial intelligence, software testing},%
colorlinks,linkcolor=darkgray,urlcolor=darkgray,citecolor=darkgray,anchorcolor=darkgray,bookmarksopen,bookmarksnumbered]{hyperref}
\usepackage[utf8]{inputenc} 
\usepackage[english,ngerman]{babel} 

\renewcommand{\orcidID}[1]{\href{https://orcid.org/#1}{%
      {\,\includegraphics[height=1.2\fontcharht\font`A]{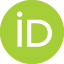}}}}

\pgfplotsset{
   compat=1.12, 
   /pgf/number format/.cd, %
      1000 sep=\thinspace, 
      min exponent for 1000 sep=4 
}

\begin{document}
\selectlanguage{english}

\title{Software Testing, AI and Robotics (STAIR) Learning Lab}
\author{
Simon Haller-Seeber\orcidID{0000-0002-1538-5906}
\and
Thomas Gatterer \and
Patrick Hofmann \and \\
Christopher Kelter \and 
Thomas Auer \and 
Michael Felderer\orcidID{0000-0003-3818-4442}
}
\authorrunning{S.~Haller--Seeber et.al.}

\institute{Department of Computer Science, University of Innsbruck \\
   Technikerstr.~21a, 6020 Innsbruck, Austria
}
\maketitle           
\vspace*{-0.3cm}
\begin{abstract}
In this paper we presented the Software Testing, AI and Robotics (STAIR) Learning Lab. STAIR is an initiative started at the University of Innsbruck to bring robotics, Artificial Intelligence (AI) and software testing into schools. In the lab physical and virtual learning units are developed in parallel and in sync with each other. Its core learning approach is based the develop of both a physical and simulated robotics environment. In both environments AI scenarios (like traffic sign recognition) are deployed and tested. We present and focus on our newly designed MiniBot that are both built on hardware which was designed for educational and research purposes as well as the simulation environment. Additionally, we describe first learning design concepts and a showcase scenario (i.e., AI-based traffic sign recognition) with different exercises which can easily be extended.

\end{abstract}
\section{Introduction}
We established our school outreach program several years ago, and apart from some Lego Mindstorms classes we saw very little robotics in Tyrolean schools. 
So we started specialized efforts to bring educational robotics and physical computing to schools (\cite{Auer-2020-ICSTW}, \cite{Haller-2020-RIE}, \cite{Lamprecht-2020-RIE}). We as well had  
specialized university courses to teach our student teachers possibilities on 
how to do robotics in schools and guide teachers to build and program robots with their students 
to participate in national  `and international robotic competitions.

In general, we want to get young people interested in science, technology, engineering and mathematics (STEM) matters, and in computer 
science in particular. We see physical computing and robotics as an intuitive and motivating 
playground -- things move and interact, making the everyday relevance of computer science tangible.
The ability to connect physical and virtual objects (things) with one another in almost any way 
has grown in importance and now permeates all areas of life, such as mobility 
(``smart mobility'', ``autonomous driving'', ``smart city''), housing (``smart home''), agriculture 
(``smart agriculture''), health (``smart health'') or production (``Industry 4.0''). 
The high practical importance and the almost unlimited possibilities to realize digital ideas and 
products also with AI technologies via e.g.~Internet of Things (IoT) systems are in contrast to the 
high level of technical understanding which is required to grasp these technologies.
In order to master the complexity of such systems and to exploit their potential,
basic understanding and basic knowledge of the underlying technologies are therefore an indispensable 
component of an innovation-oriented digital education.
The physical computing part additionally can help in forming students' knowledge and thinking and 
provide better understanding of the effects of mechanics and mathematics via not only pure 
algorithmic thinking.
Problem solving methods are developed, which give pupils and students advantages in later everyday 
life while laying a foundation for lifelong learning. 
One of the ways of making digital education more relevant is to integrate such elements of modern 
technology into appropriate teaching subjects. 
There are of course initiatives which already implemented such efforts at different educational 
levels, from school programs to small workshops, but there is very little to none in the Tyrolean area. 
Additionally, we want to widen our aims compared to our previous initiatives, therefore we further 
developed our efforts, started a new collaboration with the \emph{Media Inclusion AI Lab} were the outcome is now our joint Learning Space as part of the INNALP Education Hub\footnote{\url{https://projekte.ffg.at/projekt/4119035}}.

The Learning Lab serves the following objectives and goals in the fields of AI, robotics and software testing:
\begin{itemize}
    \item Activity-based teaching of basic and advanced knowledge in those fields
    \item Provision and develop learning materials with the involvement of teachers 
        for all ages and for any prior knowledge
    \item Provide a simulated and physical lab infrastructure to support advanced testing and teaching scenarios
\end{itemize}

\section{Hardware and Software\label{sec:minibot}}

To give students a good introduction to AI, robotics and software testing, it is important 
to provide them also with hands-on experience. This can be done in a virtual or a physical environment. 
To address both options, we do not only look at the hardware capabilities of a physical platform,
but also on the available software stack.
A physical platform itself should be affordable, easy to use for all ages, robust, and yet be 
able to showcase research or more complex algorithms. At the moment there is no suitable, off-the-shelf 
product out there. Either they are really cheap and do not provide any sensory feedback, or they are industrial 
or research products which are too expensive, the intended usage is too narrow, or they are unreliable. \cite{Karalekas-2020}, 
\cite{Haller-Seeber-2020-unp}

Therefore a combination of consumer products building up a robot which is capable of performing interesting tasks, 
ranging from moving around, self-localisation and mapping (SLAM), planning, grasping, object recognition and the ability to use modern state of the art machine learning algorithms, is attractive.

We conducted a market research and compared them to identify feasible partial solutions to build such a robot.

As a mobile base platform the choice fell on \href{https://sphero.com/products/rvr}{Sphero RVR}\footnote{
\url{https://sphero.com/products/rvr}}. 
It has a broad software support (Python, C++, etc.), the hardware is build in a stable way, includes a vast variety of 
sensors: a 9-axis IMU (3-axis gyroscope, 3-axis accelerometer and 3-axis magnetometer/compass), an rgb color sensor, a 
distance and light sensor, and it is not too expensive.

\begin{figure}
\begin{center}
    \includegraphics[trim=10 10 0 10,clip,width=\textwidth]{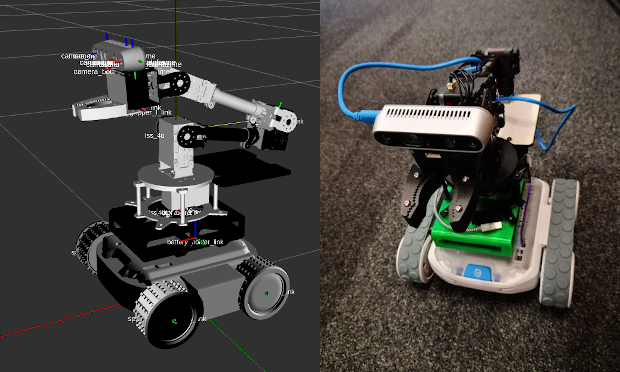}\\
    \caption{Sphero RVR with Lynxmotion Arm and in Arm Camera. (Left: in simulation. Right: as physical platform)}
    \label{minibot-hw}
\end{center}
\end{figure}

For grasping and manipulation we chose the \href{https://www.robotshop.com/de/de/lynxmotion-lss-4-dof-roboterarm-kit.html}{4 DoF
Lynxmotion Arm}\footnote{\url{https://robotshop.com/de/de/lynxmotion-lss-4-dof-roboterarm-kit.html}}.
The LSS Motors provide additional sensor feedback (voltage, current, temperature,
position, and rotation per minute) and they can set properties such as angular stiffness, holding 
stiffness, acceleration and deceleration. The software support was not too broad but we build upon the complete open reference. 
As a compute module we chose the  \href{https://developer.nvidia.com/embedded/jetson-nano-developer-kit}{Nvidia Jetson 
Nano}\footnote{\url{https://developer.nvidia.com/embedded/jetson-nano-developer-kit/}}
in combination with an \href{https://www.intelrealsense.com/depth-camera-d435/}{Intel D435 depth 
camera}\footnote{\url{https://www.intelrealsense.com/depth-camera-d435/}} which can be used for machine learning algorithms. A picture of the robot and its design is shown in Fig.~\ref{minibot-hw}.
Because this combination of hardware is new, we provide a standardized way for accessing this robot using a robot operating 
system (ROS) middelware. We chose ROS in this educational context because then we have a consistent way to control the simulation 
and a real robot. Additionally, we can address all age levels from beginners to university students. For beginners we will provide an 
easy block based web programming interface and a simple python API for intermediate pupils and students, advanced university students 
can dive into the controller back-end similar to \cite{gervais-developing-2021}.
We provide a full ROS integration for all parts and the complete robot including the following: 
\begin{description}
    \item[Nvidia Jetson:] An updated way to build an Ubuntu 20.04 image including ROS Noetic. This is important for the possibility to use 
    python3, ROS and current implementation of machine learning algorithms. We also provide an Package repository 
      for easy OpenCV and Tensorflow installation.
    \item[Sphero RVR:] We provide a full integration for the base platform, this includes especially: A ROS Description Package (including a URDF model), and a basic ROS node for robot interaction and simulation. Robot control, tf, odometry and other sensor data provisioning is done in such a way that one can benefit from the off-the-shelf ROS ecosystem (see Fig.~\ref{fig:sphero_rvr_ros}). 
    \item \begin{figure}
        \centering
        \includegraphics[width=0.9\textwidth]{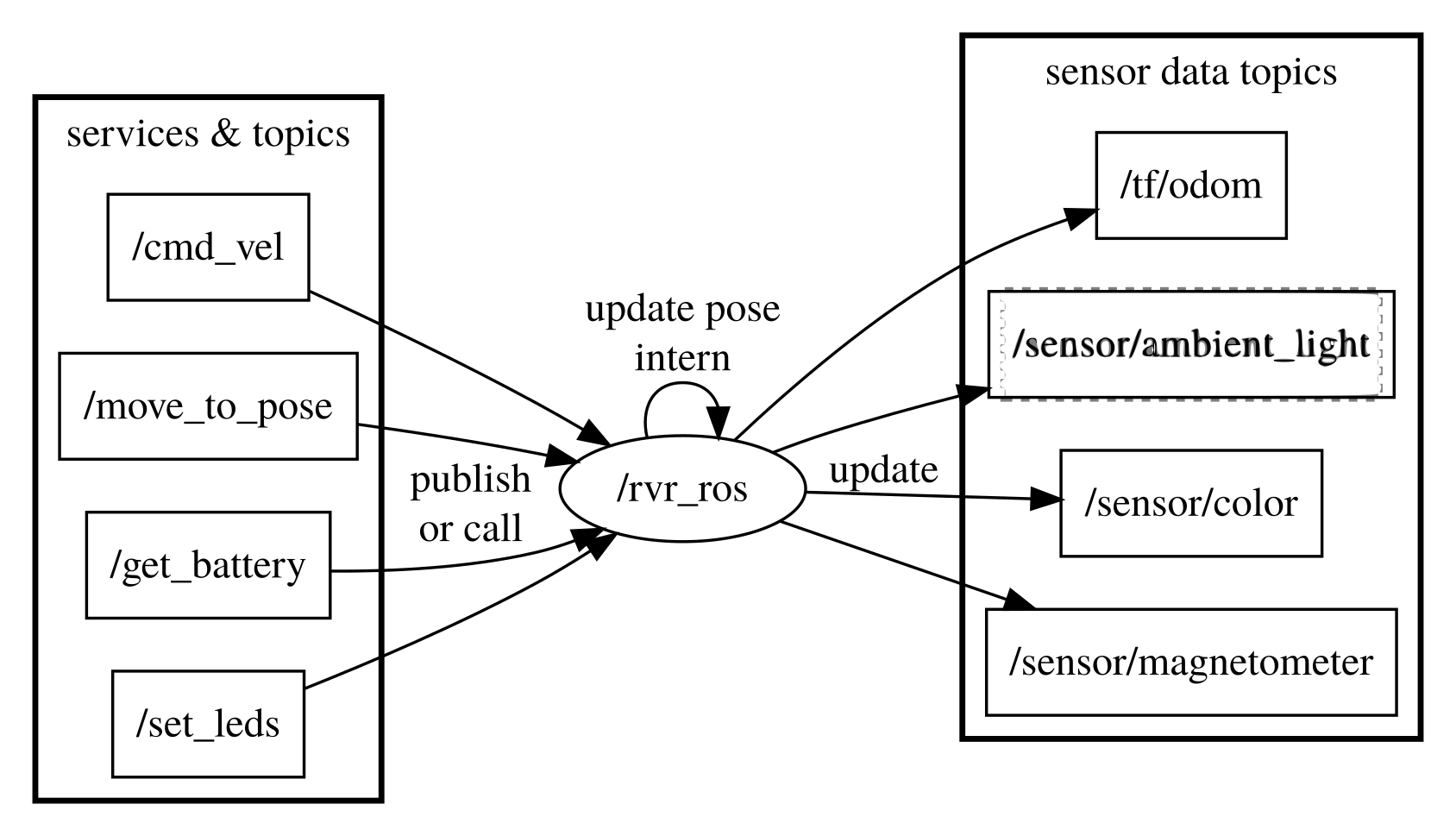}
        \caption{Sphero RVR ROS Services and Topics}
        \label{fig:sphero_rvr_ros}
    \end{figure}
    \item[Lynxmotion LSS 4DoF Arm:] We provide a full integration for this hardware, which includes: A ROS Description Package (including a URDF model), a basic ROS node for arm 
    interaction and simulation, and a ROS Arm MoveIt package for planning. A simplified view on the current ROS integration of the Arm is shown in Fig.~\ref{fig:lynxmotion_lss_ros}.
        \item \begin{figure}
        \centering
        \includegraphics[width=0.9\textwidth]{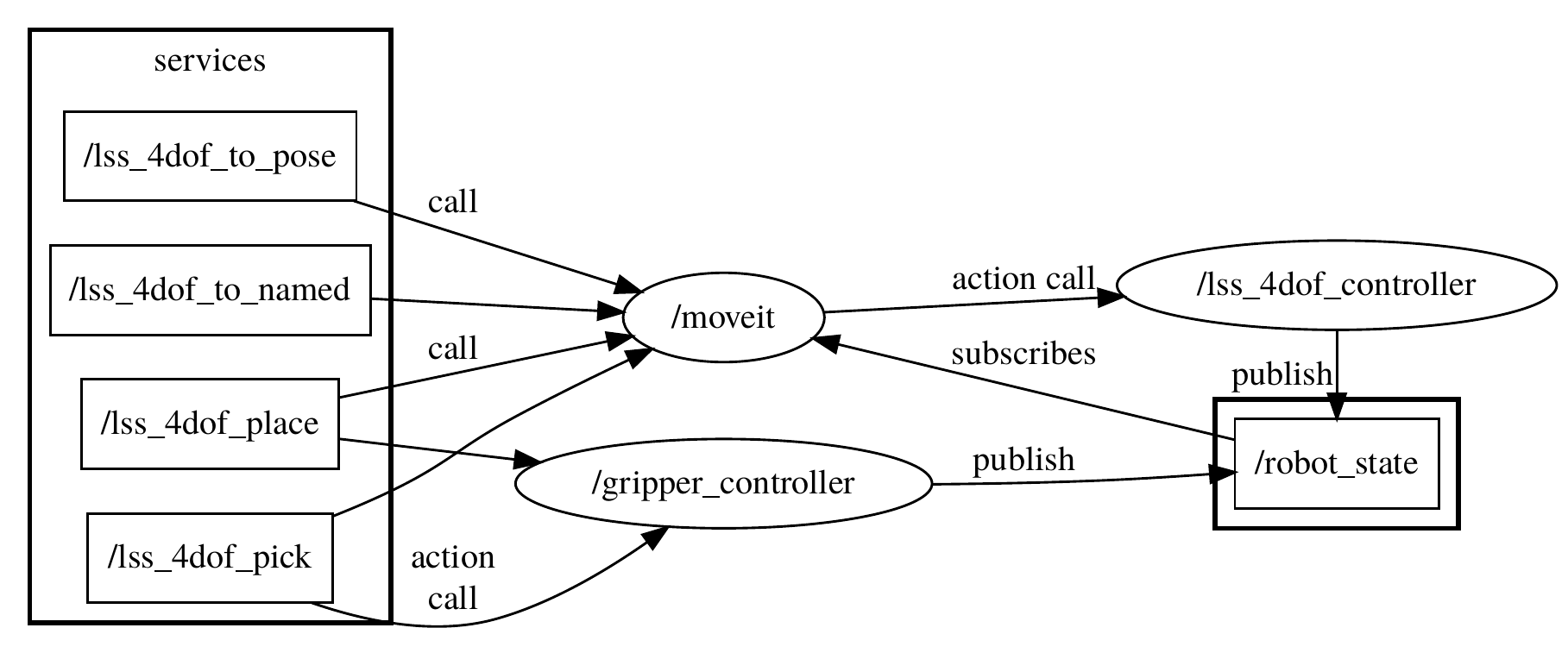}
        \caption{Simplified view: ROS integration (Services and Topics) for the Arm and Gripper}
        \label{fig:lynxmotion_lss_ros}
    \end{figure}
    \item[Intel Camera/AI:] Demos and usage examples.
\end{description}
Tips and instructions on how to build the robot hardware, as well as how to install and build the software 
setup can be found at \url{https://stair-lab.uibk.ac.at/}. The code base and implementations for the robot base and arm as well as the build environment for the Nivida Jetson can be found at: \url{https://git.uibk.ac.at/informatik/stair/}.

\section{The Learning Lab on a Running Example}
In our learning environment, we use a digital twin, among other things. This is the virtual representation of our robot and scenario. 
We use a physical simulation to not only experiment and learn, but also to evaluate exercises.
While our digital twin does not control any hardware, it can provide a valuable learning environment outside of our workshops. 
It offers an easy way to check trained AI models and written software. Additionally, we can do best practice software testing. We run, test and debug all ROS nodes either with a plugin for Visual Studio Code or with a plugin for IntelliJ IDEA.\footnote{ROS VS Code Plugin: \url{https://marketplace.visualstudio.com/items?itemName=ms-iot.vscode-ros}\\ IntelliJ Idea Plugin: \url{https://plugins.jetbrains.com/plugin/11235-ros-support}} Both work with multiple test frameworks out of the box.

In general one can use our environment with many already existing exercises and tasks from other robots and environments: From easy and beginner level block programming e.g.~building a line- or wall-follower, to intermediate ones using our simple python API (which uses our ROS modules): e.g.~exploring a maze.
In the following section we show-case a scenario which is not very common for beginner level students and briefly present possible extensions for intermediate and advanced learners.

\subsection{Traffic Sign Recognition}

\begin{figure}
\begin{center}
    \includegraphics[trim=200 200 250 100,clip,height=5.1cm]{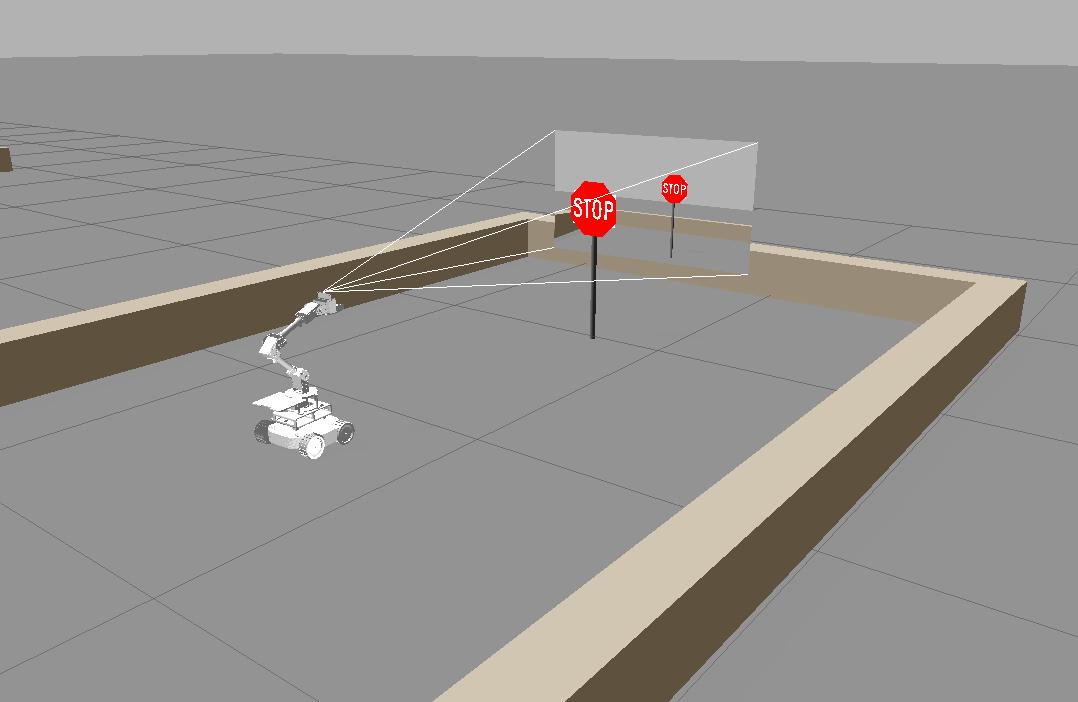}
    \includegraphics[height=5.1cm, width=4cm]{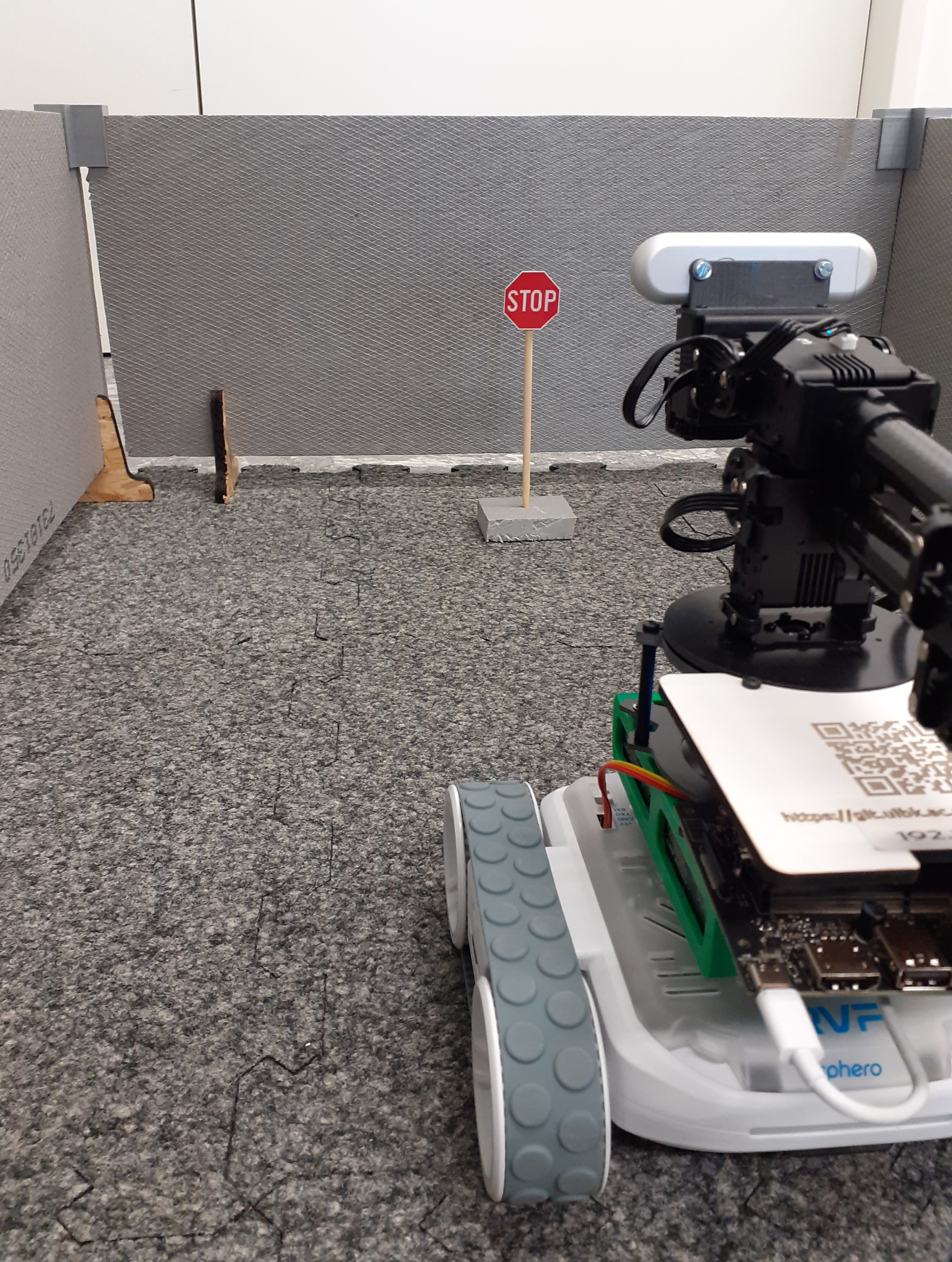}
    \caption{Our Robot in the traffic sign recognition example scenario (Left: simulated environment; Right: robot in the arena)}
     \label{fig:traffics_sign_scenario}
\end{center}
\end{figure}
In the living environment of pupils taking the driver's license test at high-school age, traffic-sign recognition is quite important. They 
recognize that the car market is changing and that autonomous cars will take over much of the content they focus on. This gives students 
a good understanding of what tasks an artificial intelligence has to do and what problems arise. After hands-on training, students can 
see the robot driving around in a real or simulated world, looking for traffic signs with its camera. Each time a traffic sign is detected, 
the image processing algorithm sends a command to the robot telling it how to respond to that sign. For example, if the robot finds a ``stop'' sign, it stops in front of the sign not continuing or going in another direction.
A visualisation of the simulated environment and in a real arena is shown in Fig.~\ref{fig:traffics_sign_scenario}.

In this scenario we have prepared three different exercises for students of different age groups and with different prior knowledge:
\begin{description}
    \item[Beginner:] Train a model with printed and/or self designed traffic signs at \href{https://teachablemachine.withgoogle.com/train/image}{Teachable 
    Machine}\footnote{\url{https://teachablemachine.withgoogle.com/train/image}}. This model can then be easily incorporated via our 
    ROS Blockly environment (see Fig.~\ref{fig:example-block}). Afterwards one has to specify actions for recognized classes in the robot control loop. Because we use ROS it is
    easy to switch between execution on the real robot and execution on the digital twin.
    \item \begin{figure}
      \begin{center}
        \includegraphics[width=\textwidth]{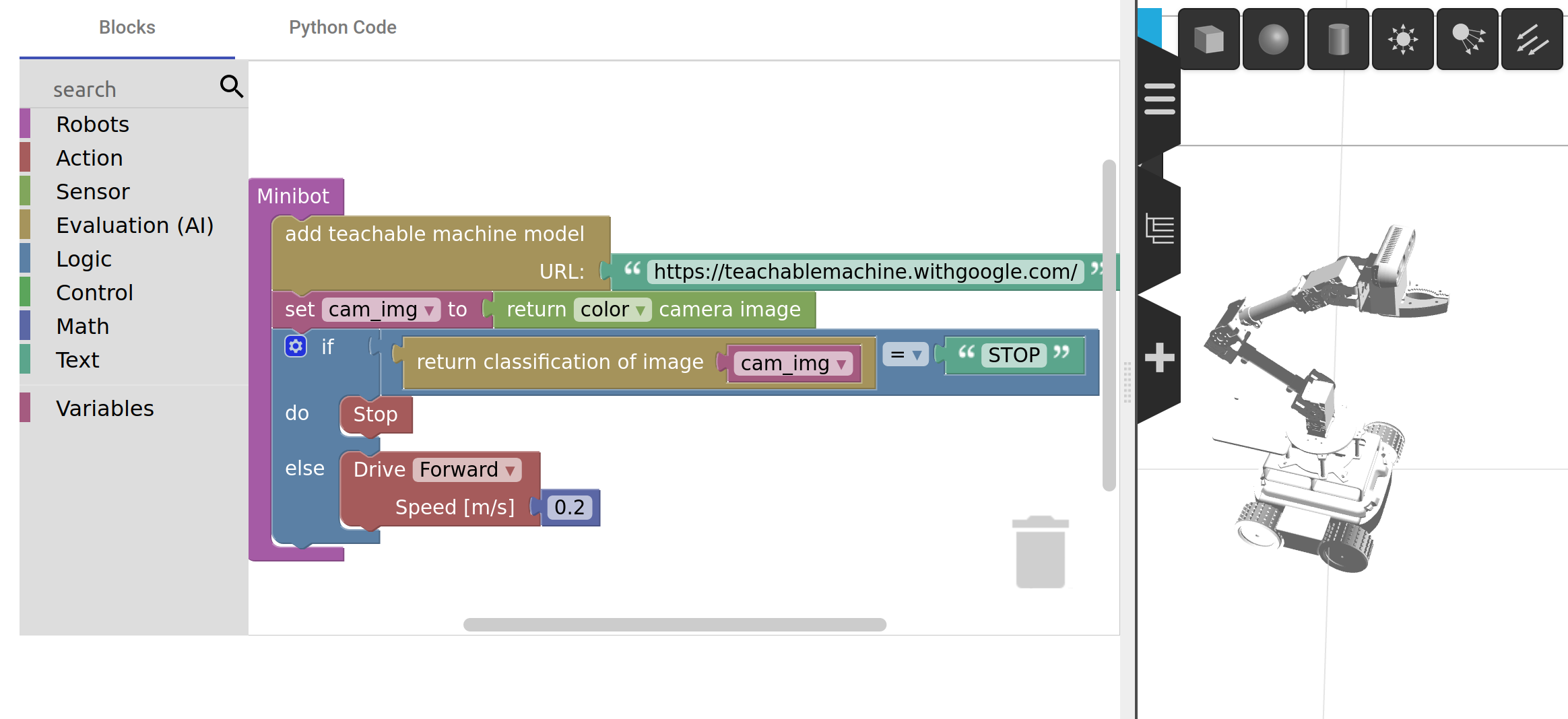}\\
        \includegraphics[width=\textwidth]{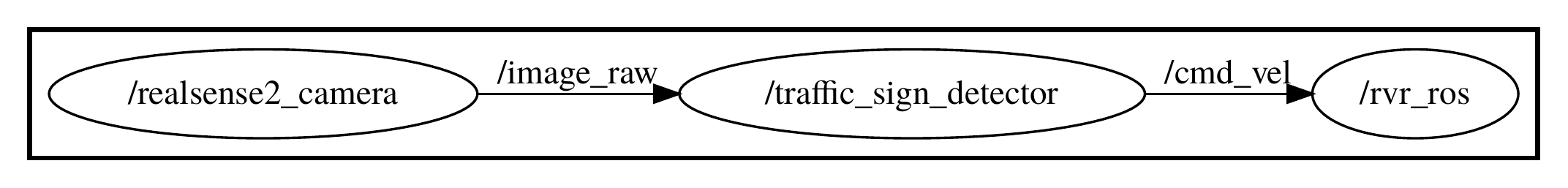}
        \caption{Left: Blocktype Programming with AI: Example program using a trained model from Teachable Machine; Right: Simulated Robot; Below: Backend - ROS nodes used for this example in our hardware environment}
        \label{fig:example-block}
      \end{center}
      \vspace*{-0.5cm}
    \end{figure}   
    \item[Intermediate:] Train a support vector machine (SVM). Compare the outcome between the beginner and this exercise. This is an interesting task, because one needs just a few training examples and low compute requirements, meaning that this could be done on the Nvidia Jetson Nano.
    \item[Advanced:] For a more advanced machine learning algorithm exercise we propose to train a YOLO model and test it. The training can't be (realistically) done on the Nvidia Jetson Nano, but it is possible to evaluate this state of the art algorithm on the robot.
\end{description}


\section{Conclusion}

In this paper we presented the Software Testing, AI and Robotics (STAIR) Learning Lab. Its core idea is the develop of a physical and simulated robotics environment in parallel and in sync with each other. In both environments AI scenarios (like traffic sign recognition) are deployed and tested.

We managed to build a robot which can be used for various courses in the fields of robotics, AI and software testing. Currently we build one scenario with different exercises for different learning environments. These can be used in simulation environment and on a physical robot. We contributed a complete ROS interface for the Sphero RVR platform, Lynxmotion LSS 4DoF Arm. Additionally, we provide the kinematics of the robot and several examples. As a next development step of our learning lab we will provide easily-accessible courses and workshops around this robot which will then be offered to Tyrolean schools in the upcoming term. In parallel, we will evaluate the workshops and their learning effects to finally provide evidence-based learning units.

\section*{Acknowledgement}
The authors want to thank the \href{https://www.uibk.ac.at/informatik/forschung/gmar-robotics-school-2021.html}{GMAR summer school participants 2021} for their valuable input. They tested an alpha version of the robots hard- and software and provided suggestions for software improvement. 
The authors also want to thank Theo Hug and Justus Piater for proofreading the manuscript. Additional financial support was provided by the Austrian Research Promotion Agency (FFG) under the scope of the research project \emph{\href{https://projekte.ffg.at/projekt/4119035}{INN\-ALP Education Hub}} (FFG contract number 4119035).

\bibliography{ref}

\begin{thebibliography}{1}
\providecommand{\url}[1]{\texttt{#1}}
\providecommand{\urlprefix}{URL }
\providecommand{\doi}[1]{https://doi.org/#1}

\bibitem{Auer-2020-ICSTW}
Auer, T., Felderer, M.: Towards a learning environment for internet of things
  testing with lego® mindstorms®. In: 2020 IEEE International Conference on
  Software Testing, Verification and Validation Workshops (ICSTW). pp. 457--460
  (2020). \doi{10.1109/ICSTW50294.2020.00081}

\bibitem{gervais-developing-2021}
Gervais, O., Patrosio, T.: {Developing an Introduction to ROS and Gazebo
  Through the LEGO SPIKE Prime}. In: Robotics in {Education}. pp. 201--209.
  Springer, Cham (Apr 2021). \doi{{10.1007/978-3-030-82544-7\_19}}

\bibitem{Haller-2020-RIE}
Haller-Seeber, S., Renaudo, E., Zech, P., Westreicher, F., Walzth\"{o}ni, M.,
  Vidovic, C., Piater, J.: {ROSSINI: RobOt kidS deSIgn thiNkIng}. In: {Robotics
  in Education}. vol.~1, pp. 16--25. Springer Advances in Intelligent Systems
  and Computing (01 2021). \doi{10.1007/978-3-030-67411-3\_2}

\bibitem{Karalekas-2020}
Karalekas, G., Vologiannidis, S., Kalomiros, J.: Europa: A case study for
  teaching sensors, data acquisition and robotics via a ros-based educational
  robot. Sensors  \textbf{20}(9) (2020). \doi{10.3390/s20092469},
  \url{https://www.mdpi.com/1424-8220/20/9/2469}

\bibitem{Lamprecht-2020-RIE}
Lamprecht, P., Haller-Seeber, S., Piater, J.: {A Block--based IDE Extension for
  the ESP32}. In: {Robotics in Education}. vol.~1, pp. 304--310. Springer
  Advances in Intelligent Systems and Computing (01 2021).
  \doi{10.1007/978-3-030-67411-3\_27}

\bibitem{Haller-Seeber-2020-unp}
{Webresource, part of University of Innsbruck course LV703051/19}: Mini
  overview on affordable physical computing platfroms which are suitable to
  build educational robots (2020),
  \url{{https://docs.google.com/spreadsheets/d/18ftAGsW0e-4IHu9VnFUISlkZ3Pj3ZBPwdiHq7ZgXySw}}

\end{thebibliography}

\end{document}